%% file: main.tex
\documentclass[runningheads, envcountsame, a4paper]{llncs}
\usepackage[T1]{fontenc}
\usepackage{array}
\usepackage{float}
\usepackage{amsmath}
\usepackage{amssymb}
\usepackage[pdftex]{graphicx}
\usepackage[caption=false]{subfig}
\usepackage{epstopdf}
\usepackage[hyphens]{url}
\usepackage[linesnumbered,ruled,vlined]{algorithm2e}

\SetKwInput{KwRequire}{Require}                
\SetKwInput{KwInput}{Input}                
\SetKwInput{KwOutput}{Output}              

\global\long\def\Ebb{\mathbb{E}}%
\global\long\def\Vbb{\mathbb{V}}%

\global\long\def\Dcal{{\cal D}}%
\global\long\def\Lcal{{\cal L}}%
\global\long\def\Ncal{\mathcal{N}}%
\global\long\def\KLD{D_{KL}}%

\begin{document}
\title{Fast Conditional Network Compression Using Bayesian HyperNetworks}
\toctitle{Fast Conditional Network Compression Using Bayesian HyperNetworks}
\author{Phuoc Nguyen \and
Truyen Tran \and
Ky Le \and
Sunil Gupta \and
Santu Rana \and
Dang Nguyen \and
Trong Nguyen \and
Shannon Ryan \and
Svetha Venkatesh}
\authorrunning{P. Nguyen et al.}
\tocauthor{Phuoc~Nguyen, Truyen~Tran, Ky~Le, Sunil~Gupta, Santu~Rana, Dang~Nguyen, Trong~Nguyen, Shannon~Ryan, Svetha~Venkatesh}
\institute{A2I2, Deakin University, Geelong, Australia \\
\email{\{phuoc.nguyen, truyen.tran, k.le, sunil.gupta, santu.rana, d.nguyen, trong.nguyen, shannon.ryan, svetha.venkatesh\}@deakin.edu.au} }

\maketitle
\setcounter{footnote}{0}

\begin{abstract}
\input{abs.tex}


\end{abstract}

\section{Introduction}

\input{intro.tex}

\section{Related Work}

\input{related.tex}

\section{Preliminaries \label{sec:Preliminaries}}

\input{prelim.tex}

\section{Conditional Compression}

\input{method.tex}

\section{Experiments}

\input{exp.tex}

\section{Conclusion}

\input{conclude.tex}

\section*{Acknowledgments}

This research was a collaboration between the Commonwealth Australia (represented by Department of Defence) and Deakin University, through a Defence Science Partnerships agreement.

\bibliographystyle{plain}
\bibliography{compression}

\end{document}

%% file: abs.tex
We introduce a conditional compression problem and propose a fast
framework for tackling it. The problem is how to quickly compress
a pretrained large neural network into optimal smaller networks given
target contexts, e.g. a context involving only a subset of classes
or a context where only limited compute resource is available. To solve
this, we propose an efficient Bayesian framework to compress a given
large network into much smaller size tailored to meet each contextual
requirement. We employ a hypernetwork to parameterize the posterior
distribution of weights given conditional inputs and minimize a variational
objective of this Bayesian neural network. To further reduce the network
sizes, we propose a new input-output group sparsity factorization
of weights to encourage more sparseness in the generated weights.
Our methods can quickly generate compressed networks with significantly
smaller sizes than baseline methods.

%% file: intro.tex
Modern deep neural networks used in computer vision and natural language processing are very large function approximators with millions of trainable parameters. Their flexibility has led to undesirable properties such as overparameterization \cite{fang2021mathematical} and memorization of random patterns \cite{arpit2017closer}. The large size makes it difficult to run the networks on edge devices, and wastes computational resources when run on servers. While it is sometimes possible to handcraft or deliberately search for more compact networks, it is often easier for practitioners to take large proven networks and compress them to fit the hardware and time constraints.

Recent efforts in network compression \cite{han2015deep,louizos2017bayesian,molchanov2017variational,louizos2018learning} have achieved significant progress in reducing large neural networks down to smaller sizes and cutting down run-time cost.  The most popular approaches are to prune hidden units and filters of dense layers and convolutional filters or to quantize weights into lower bit precision. While the former helps slim down a network, thus reducing its size and total floating point operations (FLOPS), the latter can make these networks run on limited compute hardware with integer arithmetic \cite{jacob2018quantization}. However, there is a compression-accuracy trade-off that prevent compressing large networks further and hinder their deployment in certain contexts with very low compute resources.  It is largely unknown how to determine \emph{a priori} if a certain degree of compression is achievable while satisfying context-specific constraints.

Addressing this limitation we introduce \emph{the problem of contextual compression where a large network is conditionally compressed given operating constraints}. For instance, we need to deploy a large object recognition network trained on ImageNet with 1,000 classes in scenarios requiring only a subset of the classes. Using a single compressed network is clearly wasteful, and compressed networks are likely to bias minority classes \cite{hooker2020characterising}. Thus, there is an urgent need to optimally compress networks for different contexts.

Using only a partial network at inference time has been studied in a recent line of work known as conditional computation \cite{bengio2013estimating,shazeer2017outrageously,bejnordi2019batch}.  The strategy here is to use gating units to turn off unnecessary filters conditioned on inputs, therefore saving computation. While this reduces FLOPS at inference time, we still need to store and load the whole architecture onto run-time memory, making it unsuitable for edge devices.

To this end we propose a novel Bayesian framework for conditional compression given contextual run-time constraints. Key to the framework is a hypernetwork to parameterize the posterior distribution of weights of the target network given conditions. Given a pretrained large network, our hypernetwork generates compressed network weights for \emph{on-the-fly} for each context. The conditional compression is implemented during training phase by minimizing a variational objective of the generated Bayesian neural network under different contexts. By using a sparsity inducing priors such as the scale mixture prior family \cite{louizos2017bayesian} we can force the hypernetwork to generate very sparse hidden units networks, thus implementing context-based pruning. To encourage sparsifying the input and output neurons of layers jointly, we propose a novel input-output group sparsity factorization of the weight posterior distribution, in which the (input) group sparsity factorization of \cite{louizos2017bayesian} is a special case. The framework is dubbed Bayesian Hypernetwork Compression (BHC).

We investigate the power of BHC on three general conditioning contexts: (a) given the subset of input features, (b) given the subset of output classes, and (c) constraints on expected model size. The model architectures include the popular LeNet-300-100 and LeNet-5-Caffe on MNIST dataset, and the VGG on CIFAR 10. Through a suite of experiments we demonstrate that BHC can quickly compress networks with sparse connectivity and low bit precision, resulting in significantly smaller memory footprints and computation than competing methods.

%% file: related.tex
Closest to our work are conditional computation methods \cite{bengio2013estimating,chen2018gaternet,gao2018dynamic,bejnordi2019batch}
which can predict and turn off unnecessary filters before running,
thus saving computation at inference time. However, these methods
do not work for small compute devices. It also takes an extra step
for the computation of which filters to include or remove. In \cite{chen2018gaternet}
the masks for activation of each hidden layers are computed in parallel
and for each sample independently, whereas in \cite{yang2019condconv}
the convolution kernels are parameterized using a condition from each input sample.
In batch-shaping \cite{bejnordi2019batch}, large networks are slimmed
down by using a residual network to predict masks. This is in contrast
to our contextual compression proposal where a compressed network
is generated once per context for run-time inference of any amount
of data.

Also related to ours are meta-learning based compression methods.
In \cite{li2020dhp}, a hypernetwork is used to generate pruned convolution
filters from sparse embedding vectors. In \cite{liu2019metapruning},
a similar hypernetwork is used to generate various weights for a target
network, then an evolutionary procedure is used to search for good-performing
pruned networks. In contrast, we introduce a fast contextual compression
problem where compressed networks is immediately generated given input
conditions. We develop a principled Bayesian framework to model sparse posterior
of weights which facilitates both pruning and bit precision reduction.

%% file: prelim.tex
\subsection{Bayesian neural networks \label{subsec:Bayesian-neural-networks}}

A neural network is fully specified its computational graph whose
nodes represent neural units equipped with activation functions and
connectivity is specified by weight matrices. Let $W$ be the model weights.
Bayesian neural network \cite{neal2012bayesian} provides a principled
way of modeling uncertainties over weights $W$. By choosing a suitable
prior distribution $p(W)$, we can use Bayes formula to update the
posterior distribution of weights given data $\Dcal$ as $p(W\mid\Dcal)=\frac{p(\Dcal\mid W)p(W)}{p(D)}$.
This direct Bayesian inference is a long standing problem where the
normalization term in the denominator requires marginalizing over
the entire parameter space and is intractable to compute. Many recent
advances in approximate inference for neural networks can be classified
into sampling based or variational \cite{blei2017variational}. While
the former can give better posterior approximation, recent literature
focuses on the latter for neural networks due to its scalability \cite{kingma2013auto,rezende2014stochastic,kingma2015variational}.
Variational inference for the weight distribution often uses a
simple posterior distribution such as normal distribution,
$W\sim q(W)=\Ncal(\mu_{W},\Sigma_{W})$ \cite{neal2012bayesian,louizos2017bayesian}.
Inference is then amounted to optimizing its parameters,
\[
\max_{\mu_{W},\Sigma_{w}}\Ebb_{W\sim q(W|\mu_{W},\Sigma_{W})}p(\Dcal|W)-\KLD\left(q(W|\mu_{W},\Sigma_{W})\|p(W)\right)
\]
An emerging problem for training Bayesian deep networks with many layers is in mini-batch
training to optimize its posterior distribution. While reparameterization
trick \cite{kingma2013auto,rezende2014stochastic} can be used to
sample one set of weights per update, it requires independent
weights for independent inputs in each data mini-batch. Authors in
\cite{kingma2015variational} observed that by factorizing weights
and noise and grouping noise by input neurons, they can ``apply'' the noise into the input instead of weights.
Then the (preactivation) output of the layer will be a normal distribution, linear transformation of the normal input
vector \cite{kingma2015variational},
\begin{align*}
y & =x(Z \odot W)=(x \odot z)W\\
y_j & \sim\Ncal(\sum_i w_{ij}z_ix_{i},\sum_i w_{ij}^2z_i^2x_i^2)
\end{align*}
where the noise vector $z$ is reparameterized from the noise matrix $Z$ by grouping by input neurons, $x$ is the input vector, and $y=(y_j)$ is the output vector of the layer. This interpretation is used in variational
dropout \cite{molchanov2017variational} for sparsifying networks.

\subsection{Bayesian compression \label{subsec:Bayesian-compression}}

In \cite{louizos2017bayesian}, the authors explicitly describe the
sparsity encouraging prior for neural network parameters by using
a scale mixture prior from Bayesian statistics. Thereby, minimizing
the KL divergence from this sparse prior to the weight posterior distribution
will force weights to be sparse. This Bayesian compression framework
for deep learning can be stated as follows.

Let us consider a particular weight matrix $\left[w_{ij}\right]_{i\in\overline{1,I};j\in\overline{1,J}}$
, where $w_{ij}$ is the connection from the $i$-th input unit to
the $j$-th output unit, for $I$ input units and $J$ output units.
We define a scale mixtures of normals for the prior over parameter
$w$,
\[
w_{ij}\sim\Ncal(w_{ij};0,z_{ij}^{2})
\]
where $z\sim p(z)$ is some sparse prior. Marginalizing over $z$
gives heavier tails as well as peaks at zero, thus resulting in a
sparse network.  Instead of using one scale for each weight, it is
beneficial to use one scale $z_{i}$ for all connections from neuron
$i$. This has the effect of pruning (dropping) the entire neuron.
The prior distribution of weight and scale variables are designed
as:
\begin{equation}
p(z,W)=\prod_{i=1}^{I}p(z_{i})\prod_{i,j}^{I,J}N(w_{ij}|0,z_{i}^{2})\label{eq:joint-prior}
\end{equation}
A popular choice for the scale variable is the normal-Jeffreys prior
$p(z)=\frac{1}{|z|}$. This improper prior (not normalizable) is non-informative
and non parametric which make it suitable to model the scale for weights
since it allows both small scales (weights are peak at zeros) and
large scales (weights can spread out without being shrunk). Due to
its irregularity, numerical approximation need to be made when computing
KL divergences \cite{kingma2015variational,molchanov2017variational}.
We suggest the references in \cite{louizos2017bayesian,ghosh2017model}
for more prior choices.

 The posterior of weights and scales are approximated by normal distributions
and they are factorized in the same way as the prior:
\begin{align*}
q_{\phi}(z,W) & =\prod_{i=1}^{I}N\left(z_{i}\mid\mu_{z_{i}},\sigma_{z}^{2}\right)\prod_{i,j}^{I,J}N\left(w_{ij}\mid z_{i}\mu_{ij},z_{i}^{2}\sigma_{ij}^{2}\right)
\end{align*}
Note that $\sigma_{z}^{2}$ was reparameterized as $\mu_{z_{i}}^{2}\alpha_{i}$
in \cite{wang2013fast,kingma2015variational} for the Gaussian dropout
interpretation, where $\alpha_{i}$ is the dropout rate of the neuron
$i$. This dropout rate $\alpha_{i}$ can be later compared to a threshold
to decide a dropout mask for the input neurons.

The optimization of the variational parameters $\phi=(\mu_{ij},\sigma_{ij}^{2},\mu_{z_{i}},\sigma_{z_{i}}^{2})$
is minimizing a variational lower bound of the data log-likelihood
$\log p(\mathcal{D})$:
\begin{align*}
L(\phi) & =\mathbb{E}_{q_{\phi}(z)q_{\phi}(W\mid z)}\log p(\mathcal{D}\mid W)-\mathbb{E}_{q_{\phi}(z)}\KLD\left(q_{\phi}(W\mid z)\|p(W\mid z)\right)\\
 & \qquad-\KLD\left(q_{\phi}(z)\|p(z)\right)
\end{align*}
At test time, one can recover a deterministic network by replacing
the distribution of $W$ by its posterior mean times a dropout mask.


%% file: method.tex
While the Bayesian compression framework of \cite{louizos2017bayesian},
explained in Section~\ref{subsec:Bayesian-compression}, provides
a principled way for unconditional compression, it is not clear how
to tune it to work in practical contexts with specific resource constraints.
Thus we formulate a \emph{new problem of conditional compression},
where we infer the network sparsity and bit precision \emph{on-the-fly}
conditioned on a specified context.

The compression context can be ideally arbitrary. In this work, we
focus on three general types of conditioning context: (a) the subset
of input features, (b) the subset of output classes, and (c) the expected
model size. The input feature condition means that the feature or
data distribution is limited to some subspace or input region, for
example a subset of image channels or a subset of vocabulary. The
output classes condition means that the classification output is limited
to some sub-classes. The model size constraint reflects resource specification
such as memory footprint, and this translates to reaching an equivalent
compression rate for a given pre-trained network.

We then propose a new Bayesian framework to solve the problem, employing
neural hypernetworks \cite{ha2016hypernetworks} to generate the distribution
of compressed networks given the conditions. We term this new framework
BHC, which stands for \textbf{B}ayesian \textbf{H}ypernetwork \textbf{C}ompression.

\begin{center}
\begin{figure}
\begin{centering}
\emph{\includegraphics[viewport=0bp 0bp 280bp 160bp,clip,height=4cm]{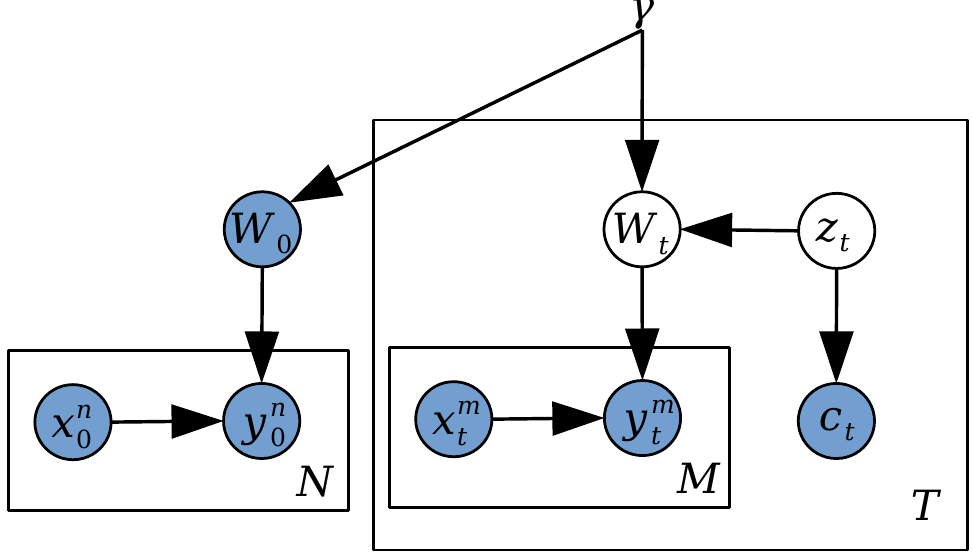}}
\par\end{centering}
\emph{\caption{The graphical model of our conditional compression framework. $W_{0}$
is the pretrained network weights trained on the dataset $\protect\Dcal_{0}=\{(x_{0}^{n},y_{0}^{n})\}_{n=1}^{N}$.
$W_{0}$ is used to initialize the hypernetwork $\gamma$ to generate
the initial means and variances of weights $W_{t}$. For each context
$t$, the hypernetwork $\gamma$ generates a compressed $W_{t}$ conditioned
on $z_{t}$ (and $c_{t}$ implicitly). Then it is used for the classification
context on \emph{$\protect\Dcal_{t}=\{(x_{t}^{m},y_{t}^{m})\}_{m=1}^{M}$}.
When$W_{0}$ is absent, the framework can also be trained from scratch
to generate $W_{t}$ for each context $t$. \label{fig:Conditional-compression-framework}}
}
\end{figure}
\par\end{center}

\subsection{Conditional Bayesian hypernetworks}
Let $W_{0}$ be the pretrained network weights trained on the dataset
$\Dcal_{0}=\{(x_{0}^{n},y_{0}^{n})\}_{n=1}^{N}$. We aim to quickly
generate a compressed network $W_{t}$ from $W_{0}$ in some context
$t$ with a constraint on the data domain or the compute resource,
denoted by the condition $c_{t}$, i.e., $W_{t}=f_{\gamma}(c_{t},W_{0})$
for some hypernetwork $f_{\gamma}$ parameterized by $\gamma$. In
theory we can train $f$ by running external compression algorithms
with varying conditions to collect training data, but this is extremely
expensive to cover the large condition space.

Inspired by the Bayesian compression framework of \cite{louizos2017bayesian}
(see Section~\ref{subsec:Bayesian-compression}), we introduce a
new efficient Bayesian framework, in that $W_{t}$ is probabilistically
generated by the hypernetwork at training time and the posterior of
$W_{t}$ from the condition $c_{t}$ is estimated at inference time.
Fig.~\ref{fig:Conditional-compression-framework} shows the graphical
model of the proposed method.

In this framework, we are given a set of $T$ classification scenarios
(or contexts) $\Dcal=\{(\Dcal_{t},c_{t})\}_{t=1}^{T}$, each containing
classification data $\Dcal_{t}=\{(x_{t}^{m},y_{t}^{m})\}_{m=1}^{M}$
and constraint $c_{t}$. Marginalizing over all condition $c_{t}$'s
then gives the posterior distribution of $W_{t}$, $q(W)=\int q(W_{t}|c_{t})p(c_{t})dc_{t}$,
which is a continuous mixture distribution with mixing density $p(c_{t})$.
The compression is done by minimizing the KL distance from a normal-Jeffreys
group sparsity prior to the weight posterior \cite{louizos2017bayesian}.
At test time, the hypernetwork generates a mask and the posterior
mean of $W_{t}$ to produce a compressed deterministic network for
classification on $\Dcal_{t}$.

We solve this problem by modeling the joint distribution between the
parameter $W$, the dropout mask $z$, and the condition $c$. First,
we assume each mask is a latent variable associated with a given condition,
and their joint distribution factors as $p(c,z)=p(c|z)p(z)$. We aims
at learning an optimal variational posterior distribution $q_{\phi}(z|c_{t})$
given this condition such that the weight $W$ is sparse and still
performs well on the (conditioned) dataset $\Dcal$. That is we maximize
the variational objective:
\begin{align}
\max_{\phi}~ & \mathbb{E}_{z\sim q_{\phi}(z|c)}\left[\log p(\Dcal\mid W,z)+\log p(c\mid z)\right]\nonumber \\
 & -\Ebb_{q_{\phi}(z)}\left[\KLD\left(q_{\phi}(W|z)\|p(W|z)\right)\right]-\KLD\left(q_{\phi}(z|c)\|p(z)\right)\label{eq:conditional-objective}
\end{align}

\subsection{Compression contexts}

For the sparsity condition that directly limits the network capacity,
we observe that varying the cutoff threshold on the logarithm of dropout
rates can produce networks with varying compression rate. The normal-Jeffreys
prior give rise to a continuum of the dropout domain in the scale
posterior which supports compression at various rate. Whereas the
horseshoe prior tends to separate the compression rates into two isolated
clusters. However, even for the normal-Jeffreys prior, we observe
that the accuracy will degrade significantly if the required compression
rate is significantly higher than its attained compression rate at
convergence. Therefore we proposed an additional level of compression
in addition to the group sparsity induced by the normal-Jeffreys prior
and use the hypernetwork to select the level of compression given
the required sparsity condition.

\subsection{Group sparsity and input-output group sparsity reparameterization}

In \cite{louizos2017bayesian}, the author proposed coupling the scales
of weights belonging to the same input neuron by sharing the scale
variable $z$ in the joint prior as well as in the joint posterior,
\begin{align*}
p(W,z) & \propto\prod_{i}^{A}\frac{1}{|z|}\prod_{i,j}^{A,B}\Ncal(w_{ij}|0,z_{i}^{2})\\
q_{\phi}(W,z) & =\prod_{i}^{A}\Ncal(z_{i}|\mu_{z_{i}},\sigma_{z_{i}}^{2})\prod_{i,j}^{A,B}\Ncal(w_{ij}|z_{i}\mu_{ij},z_{i}^{2}\sigma_{ij}^{2})
\end{align*}
This group sparsity distribution $p(z)$ is applied to the input of
a layer with weight $W$. For deep networks with multiple layers $\{W_{l}\}_{1}^{L}$,
we observe that the input sparsity $z^{l}$ of a layer $W_{l}$ should
directly affect the input sparsity of the next layers. Therefore we
introduce and additional scale variable $s$ for the output and propose
to use a joint input-output group sparsity as follow,
\begin{align*}
p(W,z,s) & \propto\prod_{i}^{A}\frac{1}{|z_{i}|}\prod_{j}^{B}\frac{1}{|s_{j}|}\prod_{i,j}^{A,B}\Ncal(w_{ij}|0,z_{i}^{2}s_{j}^{2})\\
q_{\phi}(W,z,s) & =\prod_{i}^{A}\Ncal(z_{i}|\mu_{z_{i}},\sigma_{z_{i}}^{2})\prod_{j}^{B}\Ncal(s_{j}|\mu_{s_{j}},\sigma_{s_{j}}^{2})\prod_{i,j}^{A,B}\Ncal(w_{ij}|z_{i}s_{j}\mu_{ij},z_{i}^{2}s_{j}^{2}\sigma_{ij}^{2})
\end{align*}
Note that this method applies to activation functions which have a
zero fixed point, such as tanh, RELU, and ELU families, in order for
the output group sparsity to carry on to the next layer input\footnote{We assume the bias, if exist, can be included into the weight matrix
and the input is augmented by a constant one.}. This creates a ``pairwise'' sparsity coupling one layer and the
next layer since the output feature map of one layer will be the input
to the next layer, where its sparsity pattern will be taken into account.
This coupling of the input-output sparsity directly help increase
the sparsity level of the network, thus can be used to achieve higher
compression level. In the forward step the weights $W$ is sampled
conditioned on the sparsity of both the input, via $z$, and the output,
via $s$. In the backward step, the gradient of $W$ is passed directly
to update the input and output scales in pair.

The variational lower bound under this new prior and posterior is:
\begin{align}
\Lcal(\phi) & =\Ebb_{q_{\phi}(z,s)q_{\phi}(W|z,s)}[\log p(\Dcal|W)]+\nonumber \\
 & \hspace{5mm}-\lambda_{KL}\left(\Ebb_{q_{\phi}(z,s)}[\KLD(q_{\phi}(W|z,s)\|p(W|z,s))]+\KLD(q_{\phi}(z,s)\|p(z,s))\right)\label{eq:conditional-objective-detail}
\end{align}
where $\lambda_{KL}$ is the weighting factor to control the level
of compression, which we will utilize in the experiments for different
level of compression requirements. The negative KL divergence from
the joint normal-Jeffreys scale prior to the joint Gaussian posterior
in this new parameterization is as follow:
\begin{align*}
-\KLD(q_{\phi}(z,s)\|p(z,s))\approx & \sum_{i}^{A}\left(k_{1}\sigma(k_{2}+k_{3}\log\alpha_{i})-0.5m(-\log\alpha_{i})-k_{1}\right)\\
 & +\sum_{j}^{B}\left(k_{1}\sigma(k_{2}+k_{3}\log\beta_{j})-0.5m(-\log\beta_{j})-k_{1}\right),
\end{align*}
where $\sigma(.)$, $m(.)$ are the sigmoid and softplus functions
and $k_{1}=0.63576$, $k_{2}=1.87320$, $k_{3}=1.48695$, and $\log\alpha_{i}$ and $\log\beta_{j}$ are the dropout-rate vectors of the input and output respectively.

At test time, we use the posterior mean of $W$ and apply the masks
$Z=m_{z}m_{s}^{T}$ to output the deterministic weight matrix,
\begin{equation}
\hat{W}= Z \odot\Ebb_{q_{\phi}(z,s)q_{\phi}(\tilde{W})}\left[\text{diag}(z)\tilde{W}\text{diag}(s)\right]=\text{diag}(m_{z}\odot\mu_{z})M_{w}\text{diag}(m_{s}\odot\mu_{s})\label{eq:mask}
\end{equation}
 where the mask $Z=m_{z}m_{s}^{T}$ is the binary mask matrix determined
 using the input mask $m_{z}$ and output mask $m_{s}$ determined by thresholding $z$ and $s$, and $M_w=(\mu_{ij})$ is the matrix of weight means. We use the
following variational posterior marginal variance to determine the
bit precision of the weights:

\[
\Vbb(w_{ij})=\Vbb(z_{i}\tilde{w}_{ij}s_{j})=(\sigma_{z}^{2}+\mu_{z}^{2})(\sigma_{ij}^{2}+\mu_{ij}^{2})(\sigma_{s}^{2}+\mu_{s}^{2})-\mu_{z}^{2}\mu_{ij}^{2}\mu_{s}^{2}
\]

\subsection{Training methods}

We train the Bayesian compression models then quickly adapt to each
condition using our framework. Therefore, when there is no condition,
our method generates the compressed network which matches the performance
of existing Bayesian compression networks. However when there are
extra context information, we can input into our hypernetwork to generate
a tailored, highly compressed network for that specific context.

\begin{algorithm}[H]
\DontPrintSemicolon
    \KwRequire{input $H$, hypernetwork $f_{\phi}$, condition $c\sim p(c)$}
    $M_{w},\Sigma_{w},\mu_{z},\sigma_{z},\mu_{s},\sigma_{s}=f_{\phi}(c)$ \;
    $E_{z}\sim\Ncal(0,1)$ \;
    $Z=\mu_{z}+\sigma_{z}\odot E_{z}$ \;
    $E_{s}\sim\Ncal(0,1)$ \;
    $S=\mu_{s}+\sigma_{s}\odot E_{s}$ \;
    $M_{h}= ((H \odot Z) M_{w}) \odot S $ \;
    $V_{h}= ((H \odot Z)^{2} \Sigma_{w}) \odot S^{2}$ \;
    $E\sim\Ncal(0,1)$ \;
    return $M_{h}+\sqrt{V_{h}}\odot E$ \;
\caption{BHC feedforward pass for fully connected layers. \label{alg:fc-algorithms.}}
\end{algorithm}

\begin{algorithm}[H]
\DontPrintSemicolon
    \KwRequire{input $H$, hypernetwork $f_{\phi}$, condition $c\sim p(c)$}
    $M_{w},\Sigma_{w},\mu_{z},\sigma_{z},\mu_{s},\sigma_{s}=f_{\phi}(c)$ \;
    $E_{z}\sim\Ncal(0,1)$ \;
    $Z=\mu_{z}+\sigma_{z}\odot E_{z}$ \;
    $E_{s}\sim\Ncal(0,1)$ \;
    $S=\text{reshape}(\mu_{s}+\sigma_{s}\odot E_{s},[1,1,F,N])$ \;
    $M_{h}= ((H \odot Z) * M_{w}) \odot S$ \;
    $V_{h}= ((H \odot Z)^{2} * \Sigma_{w}) \odot S^{2} $ \;
    $E\sim\Ncal(0,1)$ \;
    return $M_{h}+\sqrt{V_{h}}\odot E$ \;
\caption{BHC feedforward pass for convolution layers. \label{alg:cnn-algorithms.}}
\end{algorithm}

Algorithms~\ref{alg:fc-algorithms.} and \ref{alg:cnn-algorithms.}
show the forward pass of BHC networks for fully connected and convolution
layers respectively. In each algorithm, $M_{w}=(\mu_{ij})$ and $\Sigma_{w}=(\sigma_{ij}^2)$ are
the generated means and variances of each layers, $\mu_{z}$, $\sigma_{z}$
are the means and variances of the input group scale variables, $\mu_{s}$,
$\sigma_{s}$ are the means and variances of the input group scale
variables, $H$ is the input to the current layer, $N$ is the batch
size, $F$ is the number of convolutional filters, and $*$ is the
convolutional operator. For fully connected layers, we use one input
(output) scale variable for each input (output) neuron to group weights
connected to that neuron. For convolutional layers, we use only one
input scale variable for all filters and one output scale variable
for each filter output. We use the {[}width, height, filters, batch{]}
dimension ordering. Local reparameterization \cite{kingma2015variational}
is used to efficiently sample the input, output group scale variables,
and the activations. Both algorithms work as follow. In step 1, the mean and variance parameters of weights, input and output masks posterior distributions are generated given the condition $c$. In step 2-3, an input noise matrix $Z$ is sampled for the input matrix $H$. In step 4-5, an output noise matrix/tensor $S$ is sampled for the output matrix/tensor. In step 6 and 7, the output mean $M_h$ and variance $V_h$ is calculated and masked accordingly. Finally, an output sample is return by step 8-9.

%% file: exp.tex
\subsection{Settings}

We design the experiments with varying conditions and test whether
our Bayesian Hypernetwork Compression (BHC) method can compress and
speedup a target network tailored to each condition. We examine three
classes of conditions: \emph{conditions on the input domain, conditions
on the output classes, }and \emph{conditions on the network capacity}.

\subsubsection{Network architectures:}

We investigate our method on the well-known neural network architectures
of LeNet-300-100 and LeNet-5-Caffe on MNIST dataset, similarly with
\cite{molchanov2017variational}, and VGG on CIFAR 10, similarly with
\cite{louizos2017bayesian}. The input (output) groups of weight parameters
are designed by coupling the scale variable for each input (pre-activation
output) neuron for fully connected layers. For convolutional layers,
we use a single scale variable for the input and coupling the scale
variable for each filter. We simply use the same threshold for all
layers for pruning and vary this threshold to determine the desired
compression rate for each context\footnote{Better thresholds can be chosen by visually inspection on the log
dropout rates.}. We simply ignore the reconstruction loss of the condition, $\log p(c\mid z)$, in the optimization objective, Eq.~\ref{eq:conditional-objective},
and consider the conditions in the train and test phases come from
similar distributions to simplify the optimization problem. We leave
this investigation for future work\footnote{This involves adding a generative network for the condition posterior given all layer-masks then maximizing its log-likelihood.}.

\subsubsection{Data preparation:}

In each experiment, we divide the dataset $\Dcal_{\text{train}}$
test set $\Dcal_{\text{test}}$ into tasks (contexts), add in a condition
in each task, and finally create conditional datasets $\Dcal_{\text{train}}=\{(X_{t},Y_{t}),c_{t})\}_{t=1}^{T}$
and $\Dcal_{\text{test}}=\{(X_{t},Y_{t}),c_{t})\}_{t=1}^{T^{'}}$.
Our proposed networks are initialized with the given target network
weights, then trained in minibatchs where each batch is a different
task drawn randomly from $\Dcal_{\text{train}}$. In the test phase,
the hyper-network will generate the posterior mean over weights and
masks given the condition. A threshold can then be varied and chosen
to set the binary mask for weights such that a final compression rate
is achieved for the final deterministic weights.

\subsubsection{Baselines:}

We use the following baselines: sparse variational dropout (SparseVD) \cite{molchanov2017variational},
Bayesian compression with group normal-Jeffreys (BC-GNJ) and with
group horseshoe (BC-GHS) \cite{louizos2018learning}. We also tested
the input-output group sparsity method separately on the Lenet-300-100
architecture (denoted BC-IO-GNJ) and demonstrated that it has better
compression rate than its BC-GNJ counter part\footnote{We also tested BC-IO-GNJ on Lenet-5-Caffe and got
better performance.}. We will use the proposed input-output group sparsity method for
our BHC networks.

\subsection{Implementation}

Our method was implemented in the Flux framework \cite{Flux.jl-2018}.
We used Adam optimizer with default parameters and train our models
for 300 epochs, and found it converges better when clipping the variances
of the first layers of the networks as in \cite{louizos2017bayesian}.
Hence the first layer variance of LeNet-300-100 was clipped to $0.2^{2}$,
the first layer variance of LeNet-5-Caffe to $0.5^{2}$, and the 64
and 128 feature maps layer in VGG to $0.1^{2}$ and the 256 feature
maps layers to $0.2^{2}$.

For \emph{subclass conditioning} we use a 2-layer MLP embedding network,
each with hidden size 100 to embed the subclasses condition vector.
For \emph{input domain conditioning} we use a 2 layers convolutional
neural network each with 20 filters, kernel width 3, and zero padding
to embed the input domain condition vector.

The hypernetwork is a set of linear layers to map the embedding vectors
into the posterior weights parameters: the posterior means, log variances,
the input and output group scales. We initialize these hypernetwork
weights using Kaiming uniform function \cite{he2015delving} with
gain $0.5$. We initialize the hypernetworks such that it generates
the pretrained weight $W_{0}$ with noise initially as follows. We
set the bias of the mean-generating hypernetwork to $W_{0}$, the
bias of log-variance-generating hypernetwork to a small number drawn
from $\Ncal(-9,1e-4)$. For the scales, we set the bias of the scale-mean-generating
hypernetwork to $\Ncal(1,1e-8)$ , and the bias of its log-variance-generating
hypernetwork to a small number drawn from $\Ncal(-9,1e-4)$.

\subsection{Conditioning on output subclasses and input domain }

For the condition on output subclasses, we take the vector of Bernoulli
probabilities of the subclasses as condition, thus each dimension
contains a probability representing the appearance of a class. We
assume that the number of practical conditions is only a subset of
all possible subsets, the power set $2^{C}$ where $C$ is the number
of classes. In this experiments we limit our investigation to conditions
that are the two consecutive random classes between 1 and $C$. In
training phase, $c_{y}$ is the Bernoulli vector of length $C$ where
each component represents the probability of appearance of that class.
In test phase, the $c_{y}$ is the binary vector representing the
appearance of the classes.

For the condition on the input domain, we use minibatchs as tasks
and use features from the minibatchs as condition. We take the sample
mean of the input distribution as the condition $c_{x}$, i.e the
arithmetic mean of the input tensor along the batch dimension. In
practice, this can be the sample mean of the distribution of a representative
dataset for the testing condition.

\subsubsection{Results on compression rates}

We present the weight compression rates of our methods to the baseline
methods in Table~\ref{tab:compression-results}. We use (i) compression
by pruning input and output neurons and (ii) compression by reducing
bit precision in addition to pruning the input and output neurons.
These scenarios are practical since pruned weights can be used to
create slimmer network, where as reduced precision weights can be
exploited in future hardware \cite{wang2019haq}. We can observe that
our method can achieve higher compression rates at lower error rates
than the baselines. When condition on the subclasses, the compression
rates is highest. When exploiting the weight variance to reduce bit
precision in combination with pruning, a much higher compression rate
is achieved, $216\times$ for the LeNet-300-100 architecture, $1004\times$
for the LeNet-5-Caffe architecture, and $73\times$ for VGG architecture
in this conditional setting.
\begin{table}
\centering{}\caption{Compression results of Bayesian-hypernet compression (BHC) compared
    to sparse variational dropout (SparseVD), group normal-Jeffreys (BC-GNJ)
    and group horseshoe (BC-GHS). Results marked with
    {*} are from \cite{louizos2017bayesian}. BHC method conditionally
    compresses the target network based on the domain input condition
    $c_{x}$ or the subclass condition $c_{y}$ in each context. We report
    the mean compression rate across tasks and the overall error on the
testset. \label{tab:compression-results}}
\begin{tabular}{>{\centering}p{0.18\textwidth}>{\raggedright}p{0.15\textwidth}>{\centering}p{0.2\textwidth}>{\centering}p{0.2\textwidth}}
\hline
Model\\
Original error & Method & Pruning (Error \%) & Pruning and bit reduction (Error \%)\tabularnewline
\hline
LeNet-300-100 & SparseVD{*} & 21 (1.8) & 84 (1.8)\tabularnewline
1.6 & BC-GNJ{*} & 9 (1.8) & 36 (1.8)\tabularnewline
 & BC-GHS{*} & 9 (1.8) & 23 (1.9)\tabularnewline
\cline{2-4} \cline{3-4} \cline{4-4}
 & BC-IO-GNJ & 12 (1.8) & 40 (1.9)\tabularnewline
 & BHC-$c_{x}$ & $26$ (1.9) & $169$ ($1.9$)\tabularnewline
 & BHC-$c_{y}$ & $54$ ($1.8$) & $216$ ($1.9$)\tabularnewline
\hline
LeNet-5-Caffe & SparseVD{*} & 63 (1.0) & 228 (1.0)\tabularnewline
0.9 & BC-GNJ{*} & 108 (1.0) & 361 (1.0)\tabularnewline
 & BC-GHS{*} & 156 (1.0) & 419 (1.0)\tabularnewline
\cline{2-4} \cline{3-4} \cline{4-4}
 & BHC-$c_{x}$ & $187$ (0.2) & $608$ (0.4)\tabularnewline
 & BHC-$c_{y}$ & 286 (0.4) & 1004 (0.5)\tabularnewline
\hline
VGG & BC-GNJ{*} & 14 (8.6) & 56 (8.8)\tabularnewline
8.4 & BC-GHS{*} & 18 (9.0) & 59 (9.0)\tabularnewline
\cline{2-4} \cline{3-4} \cline{4-4}
 & BHC-$c_{x}$ & 20 (7.0) & 66 (7.4)\tabularnewline
 & BHC-$c_{y}$ & 22 (2.0) & 73 (2.8)\tabularnewline
\hline
\end{tabular}
\end{table}

\subsubsection{Compressed models}

We present the neuron group sparsity enforcing capabilities of our
methods compared to baselines in Table.~\ref{tab:pruned-architectures}.
We show the pruned architecture and bit-precision per layer. We observe
that our methods can generate very sparse input and output neurons
of the target architecture. Especially, the input layer and the first
hidden layers have significantly sparser connection and only need
as low as 5 bit precision compared to the baselines. For Lenet-5-Caffe
and VGG architecture, the first and last layers require a much smaller
number of neurons. For Lenet-5-Caffe architecture, as low as 10 bit
precision can be achieved. For VGG, the number of convolutional filters
is also reduced significantly compared to sparse variational dropout,
group normal-Jeffreys, and group horseshoe.

\subsection{Conditioning on the model compression rate}

In this section, we investigate the Bayesian-hypernet compression
method given different sparsity constraints. We compare BHC to Bayesian
compression methods using normal-Jeffreys and horseshoe priors. We
implement BHC using two broad levels of compression as condition:
(1) high accuracy compression and (2) high compression rate with moderate
accuracy drop. Finer compression rates can be chosen by varying the
threshold in each level. Therefore, we use a binary condition input
to the hypernetwork, $c_{w}=0$ for high accuracy compression and
$c_{w}=1$ for high compression rate. In this case, the condition
is discrete and BHC becomes a finite mixture model:
\[
p(W)=\int p(W|c)p(c)dc=p(W|c)p(c=0)+p(W|c)p(c=1)
\]
with mixing probabilities $p(c=0)=p(c=1)=\frac{1}{2}$. First, the
user need to inputs $c_{w}\in\{0,1\}$ into BHC hypernetwork and get
as output the posterior mean of weights $M_{w}=(\mu_{ij})$ and a dropout-rate
vector $\log\alpha$. Then the user varies a threshold $\tau$ to
create an input mask $m_z=I(\log\alpha<\tau)$ and output mask $m_s=I(\log\beta<\tau)$ for $z$ and $s$ and calculates the deterministic
weight as $\hat{W}=\text{diag}(m_z) \odot M_{w}\odot\text{diag}(m_s)$, as in Eq.~\ref{eq:mask}.
By checking the compression rate and the accuracy of $\hat{W}$ on
a validation set if it is available, the user can decide a final threshold
$\tau$ to use.

\begin{table}
\begin{centering}
\caption{Pruned architectures and bit precision of Bayesian-hypernet compression (BHC) compared to sparse variational dropout (SparseVD) \cite{molchanov2017variational}, group
normal-Jeffreys and group horseshoe \cite{louizos2018learning}. Results
marked with {*} are from \cite{louizos2017bayesian}. \label{tab:pruned-architectures}}
\par\end{centering}
\centering{}%
\begin{tabular}{>{\centering}m{0.22\textwidth}>{\raggedright}p{0.12\textwidth}>{\centering}p{0.35\textwidth}>{\centering}p{0.3\textwidth}}
\hline
Network \& size & Method & Pruned architecture & Bit-precision\tabularnewline
\hline
Lenet-300-100 & SparseVD{*} & 512-114-72 & 8-11-14\tabularnewline
784-300-100 & BC-GNJ{*} & 278-98-13 & 8-9-14\tabularnewline
 & BC-GHS{*} & 311-86-14 & 13-11-10\tabularnewline
\cline{2-4} \cline{3-4} \cline{4-4}
 & BC-IO-GNJ & 255-90-15 & 7-9-15\tabularnewline
 & BHC-$c_{x}$ & 183-50-18 & 5-8-20\tabularnewline
 & BHC-$c_{y}$ & 141-30-18 & 6-11-20\tabularnewline
\hline
Lenet-5-Caffe & SparseVD{*} & 14-19-242-131 & 13-10-8-12\tabularnewline
20-50-800-500 & BC-GNJ{*} & 8-13-88-13 & 18-10-7-9\tabularnewline
 & BC-GHS{*} & 5-10-76-16 & 10-10-14-13\tabularnewline
\cline{2-4} \cline{3-4} \cline{4-4}
 & BHC-$c_{x}$ & 4-4-64-2 & 12-10-11-11\tabularnewline
 & BHC-$c_{y}$ & 4-4-64-3 & 12-9-11-9\tabularnewline
\hline
VGG & BC-GNJ{*} & 63-64-128-128-245-155-63- \\
-26-24-20-14-12-11-11-15 & 10-10-10-10-8-8-8- \\
-5-5-5-5-5-6-7-11\tabularnewline
\small(2$\times$64)-(2$\times$128)-\\
-(3$\times$256)-(8$\times$512) & BC-GHS{*} & 51-62-125-128-228-129-38- \\
-13-9-6-5-6-6-6-20 & 11-12-9-14-10-8-5- \\
-5-6-6-6-8-11-17-10\tabularnewline
\cline{2-4} \cline{3-4} \cline{4-4}
 & BHC-$c_{x}$ & 52-63-126-128-161-71-\\
-22-19-11-10-6-7-8-8-9-13 & 12-7-5-7-6-8-9-8-\\
-11-12-12-12-14-14-26-20\tabularnewline
 & BHC-$c_{y}$ & 42-63-125-123-110-44-15-\\
-12-8-6-5-5-6-6-6-8 & 12-11-8-13-6-9-10-\\
-8-9-9-9-10-14-18-27-24\tabularnewline
\hline
\end{tabular}
\end{table}

We train BHC for $c_{w}=0$ condition using a lower KL weighting,
$\lambda_{KL}=0.1$ in the training objective (Eq.~\ref{eq:conditional-objective-detail}),
and we use a higher KL weighting, $\lambda_{KL}=1$ for $c_{w}=1$
condition. To speed up convergence, we choose a simple training strategy
with this binary condition as follows\footnote{In general, the condition can be sampled randomly from $\text{Bernoulli}(\frac{1}{2})$
during training.}. We train the BHC hypernetwork with $c_{w}=0$ condition for 100
epochs with $\lambda_{KL}=0.1$ and freeze this generated weights
in the the hypernetwork bias output, by setting the trainable flag
of the bias to false. Next, we train the BHC hypernetwork with $c_{w}=1$
condition for another 200 epochs with $\lambda_{KL}=1$ .

\subsubsection{Results}

Fig.~\ref{fig:cw} compares our Bayesian-hypernet compression (BHC)
with Bayesian compression with normal Jeffreys and horseshoe priors
by varying the threshold to output deterministic weights. As can be
observed, our methods can generate models with high accuracy at low
compression rate and with only moderate accuracy drop at high compression
rate.

\begin{center}
\emph{}
\begin{figure}
\begin{centering}
\emph{}\subfloat[LeNet-300-100.\label{fig:LeNet-300-100}]{\includegraphics[width=0.4\textwidth,height=0.3\textwidth]{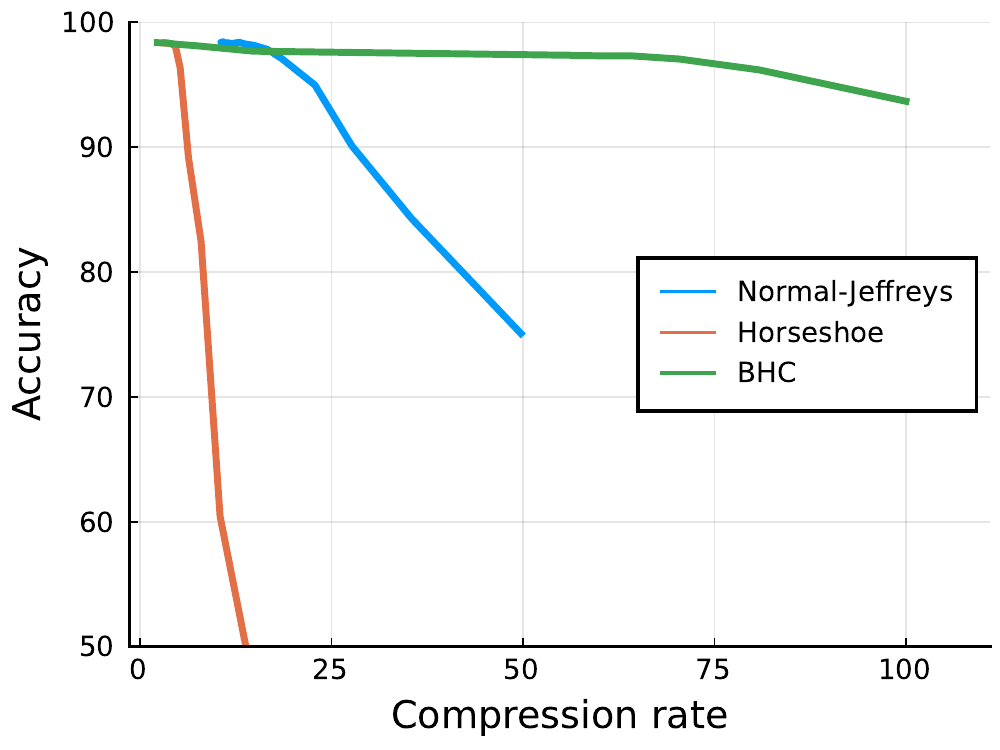}

\emph{}

}\emph{\hspace{0mm}}\subfloat[LeNet-5-Caffe.\label{fig:LeNet-5-Caffe}]{\includegraphics[width=0.5\textwidth,height=0.3\textwidth]{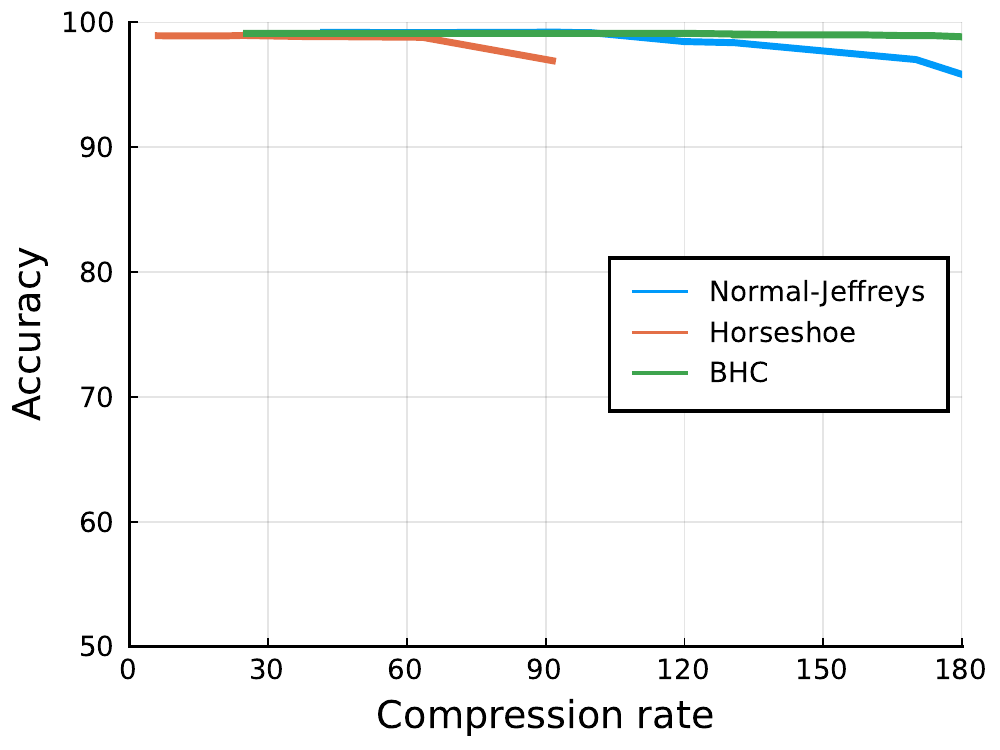}

\emph{}}\emph{}
\par\end{centering}
\emph{\caption{Conditions on compression rate. \label{fig:cw}}
}
\end{figure}
\par\end{center}

%% file: conclude.tex
We introduced a new problem concerning contextual compression of deep neural networks. Given a large uncompressed pre-trained network, we compress it \emph{on-the-fly} to satisfy resource constraints at run time, e.g., compression rate, memory, input size or class subset. Our work provides the first principled Bayesian hypernetwork compression framework towards contextual compression for edge devices.
Testing on several popular architectures on MNIST and CIFAR 10, we demonstrated that our methods can quickly generate compressed networks with  significantly smaller memory footprints and computation than competing methods.
Future challenges would be scaling up the Bayesian hypernetwork compression method to recent large architectures with hundred of millions of parameters and applying the method in real world contextual compression tasks and other data domains such as texts and sounds.